\documentclass[conference]{IEEEtran}
\usepackage{cite}
\usepackage{amsmath,amssymb,amsfonts}
\usepackage{algorithmic}
\usepackage{graphicx}
\usepackage{textcomp}
\usepackage{tikz}
\usepackage{xcolor}
\usepackage{scalefnt}
\usepackage{subcaption}
\def\BibTeX{{\rm B\kern-.05em{\sc i\kern-.025em b}\kern-.08em
    T\kern-.1667em\lower.7ex\hbox{E}\kern-.125emX}}
\usepackage{booktabs,makecell,tabularx,siunitx}
    
\usepackage{tikz}
\usetikzlibrary{backgrounds, calc, chains, fit, matrix, positioning, shadows, shapes, circuits.ee}
\tikzset{input/.style={}}
\tikzset{output/.style={}}
\tikzset{op/.style={circle, draw, thick, fill=black!10, minimum size=2.5ex, inner sep=0.1ex}}
\tikzset{filter/.style={rectangle, draw, thick, fill=black!10, minimum size=3.5ex, inner sep=1ex}}
\tikzset{other/.style={rounded rectangle, draw, fill=white, minimum size=3.5ex, inner xsep=1ex}}
\tikzset{nn/.style={trapezium, trapezium angle=80, draw, thick, fill=black!10, inner sep=1ex}}
\tikzset{branch/.style={circle, draw, thick, fill=black, minimum size=.5ex, inner sep=0ex}}
\tikzset{tensor/.style={rectangle, draw, thick, fill=white, minimum size=2em, double copy shadow={shadow xshift=.5ex,shadow yshift=-.5ex}}}
\tikzset{image/.style={rectangle, draw, thick, fill=white, minimum size=2em}}
\tikzset{block/.style={rectangle, draw, fill=white, minimum size=2em}}
\tikzset{>=direction ee}

\usepackage{pgfplots}
\pgfplotsset{compat=1.14}
\pgfplotsset{every axis/.append style={enlargelimits={abs=3pt},grid,axis lines=left}}
\pgfplotsset{every axis plot/.append style={thick,mark size=1.5pt,line join=bevel,mark options={solid}}}
\pgfplotsset{label style={font=\small}}
\pgfplotsset{tick label style={font=\footnotesize}}
\pgfplotsset{grid style={color=black!10}}
\pgfplotsset{legend style={draw=none,opacity=.85,font=\footnotesize,cells={anchor=west,opacity=1}}}
\pgfplotsset{every non boxed x axis/.style={xtick align=center,shorten <=-.5\pgflinewidth}}
\pgfplotsset{every non boxed y axis/.style={ytick align=center,shorten <=-.5\pgflinewidth}}
\pgfplotsset{every non boxed z axis/.style={ztick align=center,shorten <=-.5\pgflinewidth}}
\pgfplotsset{/pgf/number format/1000 sep={\,}}

\definecolor{gblue}{HTML}{1f77b4}
\definecolor{ggreen}{HTML}{2ca02c}
\definecolor{bgred}{HTML}{eb6c63}
\definecolor{sns1}{rgb}{0.86, 0.3712, 0.33999999999999997}
\definecolor{sns2}{rgb}{0.7247999999999999, 0.86, 0.33999999999999997}
\definecolor{sns3}{rgb}{0.33999999999999997, 0.86, 0.5792000000000002}
\definecolor{sns4}{rgb}{0.33999999999999997, 0.5167999999999995, 0.86}
\definecolor{sns5}{rgb}{0.7871999999999999, 0.33999999999999997, 0.86}

\begin{document}

\title{Fourier Basis Density Model}

\author{\IEEEauthorblockN{Alfredo De la Fuente}
\IEEEauthorblockA{\textit{Google} \\
New York, NY, 10011, USA \\
alfredodlf@google.com}
\and
\IEEEauthorblockN{Saurabh Singh}
\IEEEauthorblockA{\textit{Google Research} \\
Mountain View, CA, 94043, USA \\
saurabhsingh@google.com}
\and
\IEEEauthorblockN{Johannes Ball\'{e}}
\IEEEauthorblockA{\textit{Google Research} \\
New York, NY, 10011, USA \\
jballe@google.com}
\and
}

\maketitle

\begin{abstract}
We introduce a lightweight, flexible and end-to-end trainable probability density model parameterized by a constrained Fourier basis. We assess its performance at approximating a range of multi-modal 1D densities, which are generally difficult to fit. In comparison to the deep factorized model introduced in \cite{balle2018variational}, our model achieves a lower cross entropy at a similar computational budget. In addition, we also evaluate our method on a toy compression task, demonstrating its utility in learned compression.
\end{abstract}


\section{Introduction}
Density estimation is a ubiquitous problem in statistics and machine learning. Given a set of i.i.d. samples from an unknown true distribution, we aim to find the parameters of a density model that best describe this target distribution. In particular, the Kullback--Leibler divergence (KLD) between the model and the true distribution is often use to measure the quality of fit. Within the field of lossy neural data compression \cite{balle2020nonlinear,yang2023introduction}, the cross entropy, related to the KLD, is directly connected with the bit rate of the compression method. In this context, the model is also labeled as an \emph{entropy model}. It is critical to be able to model arbitrary density functions in order to develop efficient learned compression systems.

To fit a density model to a limited set of samples, we need to assume the density belongs to a particular class of functions. Since there may not be prior knowledge about the target distribution, restricting the model to simple parametric distributions such as Gaussians, Laplacians, etc. would not generally satisfy the need to obtain a good fit. Non-parametric approaches, which generally have extensible families of parameters, and can often guarantee fitting arbitrary functions in the asymptotic limit, includes mixtures of Gaussians \cite{bookmixture}, mixtures of Kernel functions \cite{Bunea2007SparseDE}, Parzen windows \cite{parzen1962estimation}, and others. Here, we explore the use of Fourier series to model probability densities. Any function has a Fourier series expansion, albeit with a potentially infinite sequence of coefficients. Truncating this sequence restricts the series to smooth functions, which is a reasonable implicit bias for our purposes.

A different approach based on modeling the cumulative distribution function (CDF) using a multi-layer perceptron (MLP) is the deep factorized probability (DFP) model \cite{balle2018variational}. The MLP is constrained to have strictly non-negative weights, and specialized activation functions that guarantee monotonicity of the CDF. The model has proven quite popular (e.g., \cite{minnen2018joint}), but there are questions regarding how general and parameter-efficient it is. The range of possible distributions that the DFP can model is hard to understand given the intricacies of its nonlinearities. In particular, empirical evidence suggests that the model struggles to approximate multi-modal distributions accurately.

Through a number of experiments, we analyze the properties of the proposed Fourier basis density model and how its performance can depend on the data distribution. Since neural compression models such as Nonlinear Transform Coding (NTC) \cite{balle2020nonlinear} work by compressing one dimension at a time, our proposed model is applicable for such tasks.

\section{Model Definition}
Truncated Fourier series (i.e., with all but the first $N$ coefficients assumed zero), are a canonical way to represent smooth functions. Our aim is to represent a probability density function $p(x)$ with $x \in \mathbb{R}$ as a Fourier series with a finite number of coefficients, and find these coefficients using stochastic optimization (for example, by stochastic gradient descent). In what follows, we first construct a flexible periodic density model, and then extend it to the entire real line. Note that $c^*$ and $|c|$  denote the complex conjugate and magnitude, respectively, of a complex number $c$.

Let us begin with a probability density defined as $p(x) \equiv f(x)/Z$, where $f(x)$ is a periodic (with period 2), real-valued, positive smooth function and $Z = \int_{-1}^{1}f(x)dx$ is the normalization constant. We represent $f(x)$  in terms of its complex valued truncated Fourier series coefficients $c_n \in \mathbb{C}$:
\begin{equation}
\label{eq:fourier_series}
f(x) = \sum_{n=-N}^{N} c_n \, \exp(in\pi x),
\end{equation}
where $i\equiv \sqrt{-1}$. Conversely, we can write the coefficients as
\begin{equation}
\label{eq:coefficients}
c_n = \frac 1 2 \int_{-1}^{1} f(x) \exp(-in\pi x) dx.
\end{equation}
Note that due to $f(x)$ being real-valued, the coefficients follow the symmetry $c_n = c_{-n}^\ast$ for all $n$. Consequently, the negative frequencies $n<0$ are redundant and need not be considered model parameters. Now, we desire a model that represents a flexible and valid probability density on $\mathbb{R}$, so it needs to be 1) non-negative, 2) normalized, and 3) non-periodic with the support on the full domain, $\mathbb{R}$. We ensure this as follows.

\subsection{Non-Negativity}
To guarantee non-negativity, we consider Herglotz's theorem \cite{brockwell2013time}. It states that the Fourier series of a non-negative function is positive semi-definite. In other words, $f(x)$ is non-negative if and only if $c_n$ is a positive semi-definite sequence. A simple way to ensure this is to parameterize $c_n$ as an autocorrelation sequence, i.e. for $n = 0, \ldots, N$:
\begin{equation}
\label{eq:autocorr}
c_n = \sum_{k=0}^{N-n} a_k \, a_{k+n}^\ast,
\end{equation}
where $a_n \in \mathbb{C}$ is an arbitrary sequence defined for $n=0, \ldots, N$ (and assumed zero otherwise). We can thus consider $\theta \equiv \{a_n\}_0^N$ as the parameters of the model to be fitted and guarantee, by plugging \eqref{eq:autocorr} into \eqref{eq:fourier_series}, that $f(x)$ is always non-negative.

\subsection{Normalization}
To compute the normalization constant $Z$, note in \eqref{eq:coefficients} that the integral over one period of $f(x)$ is $2c_0$. Thus, if we limit the density to a single period, the normalization constant is available directly as $Z=2c_0$. Using this normalizer, and together with non-negativity and the symmetry of $c_n$, we can now define a valid density model on $(-1, 1)$:
\begin{equation}
\label{eq:model_p}
p(x; \theta) = \frac 1 2 + \sum_{n=1}^{N} \frac {c_n} {c_0} \exp(in\pi x),
\end{equation}
where $c_n$ is as defined in Eq. \eqref{eq:autocorr}. Note that the cumulative distribution function (CDF) $P(x)$ also has a simple closed-form expression:
\begin{equation}
\label{eq:cdf}
P(x; \theta) = \frac x 2 + \sum_{n=1}^{N} \frac {c_n} {\pi in c_0} \exp(in \pi x) + C(\theta),
\end{equation}
where $C$ is a function of the parameters that ensures $P(-1) = 0$ and $P(1) = 1$.

\subsection{Support on $\mathbb R$}
Lastly, to extend this model to the entire real line, we consider the mapping $g : (-1, 1)  \rightarrow \mathbb{R}$, which is parameterized by a scaling $s$ and an offset $t$:
\begin{equation}
g(x; s, t) = s \cdot \tanh^{-1}(x) + t = \frac s 2 \ln \left( \frac{1+x}{1-x} \right) + t.
\end{equation}
The CDF $Q$ of the mapped variable can be written directly as
\begin{equation}
\label{eq:cdf_transformed}
Q(x; \theta, s, t) = P\left(g^{-1}(x; s, t); \theta\right)
\end{equation}
with
\begin{equation}
g^{-1}(x; s, t) = \tanh\left(\frac {x-t} s\right),
\end{equation}
whereas in the density $q$, the derivative of $g$ needs to be taken into account:
\begin{multline}
q(x; \theta, s, t) = p\left(g^{-1}(x; s, t); \theta\right) (g^{-1})^{'} (x; s, t) \\
= p\left(\tanh\left(\frac {x-t} s\right); \theta\right)  \frac 1 s \text{sech}^2\left(\frac {x-t} s\right).
\label{eq:model_r}
\end{multline}
  
\subsection{Weight Regularization}
In the presence of limited data, more than one choice of parameters may fit the data equally well, even for a truncated series. To express a preference towards smoother densities, we penalize the total squared variation of the unnormalized density $f(x)$ with the following regularization loss:
\begin{equation}
\label{eq:tv}
    \mathcal{L}_{\mathrm{reg}}(\theta) \equiv \gamma \int_{-1}^{1}  \bigg\lvert\frac{df(x)}{dx} \bigg\rvert^2 dx = \gamma \sum_{n=-N}^N 2\pi^2n^2 \lvert c_n \rvert^2.
\end{equation}
Here, $\gamma > 0$ is a hyperparameter specifying the weight of the regularization. This leads to an intuitive penalty on the frequency coefficients, where the coefficients for the higher frequencies are penalized more than the lower ones. The hyperparameter $\gamma$ needs to be selected manually, but in our experiments the optimization outcome is quite robust to the choice. The regularization term appears to stabilize the training dynamics as well.

To establish the equality in eq.~\eqref{eq:tv}, first note that:
\begin{align}
\frac{d f(x)}{dx} &= \sum_{n=-N}^N in\pi c_n \exp(in\pi x) \\
\bigg\lvert\frac{df(x)}{dx} \bigg\rvert^2 &= \pi^2\sum_{n=-N}^N \sum_{m=-N}^N \int_{-1}^1 nmc_n c^*_m e^{i(n-m)\pi x} dx
\end{align}
Now recall that
\begin{align}
\int_{-1}^1 \exp(i(n-m)\pi x)dx =
\begin{cases}
0 \quad \text{if}~ n \neq m \\
2 \quad \text{if}~n=m
\end{cases} 
\end{align}
which immediately leads to eq. \eqref{eq:tv}.

\begin{figure}[t]
    \centering
    \begin{subfigure}[t]{0.48\linewidth}
        \centering
        \includegraphics[width = \linewidth]{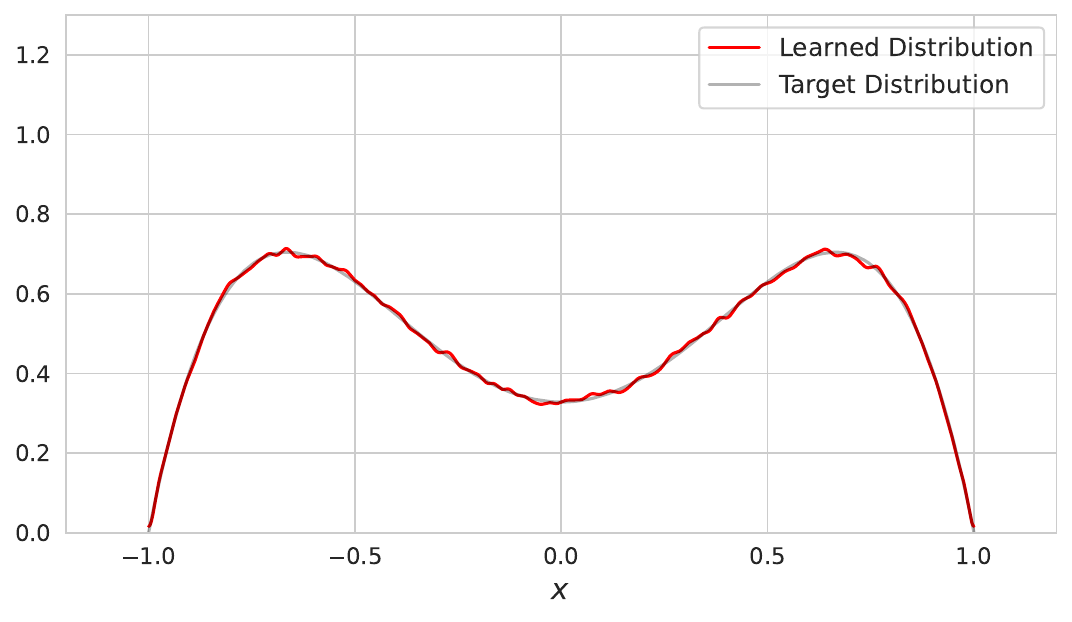}
        \caption{density fit, $N=64$}
    \end{subfigure}%
    \begin{subfigure}[t]{0.48\linewidth}
        \centering
        \includegraphics[width = \linewidth]{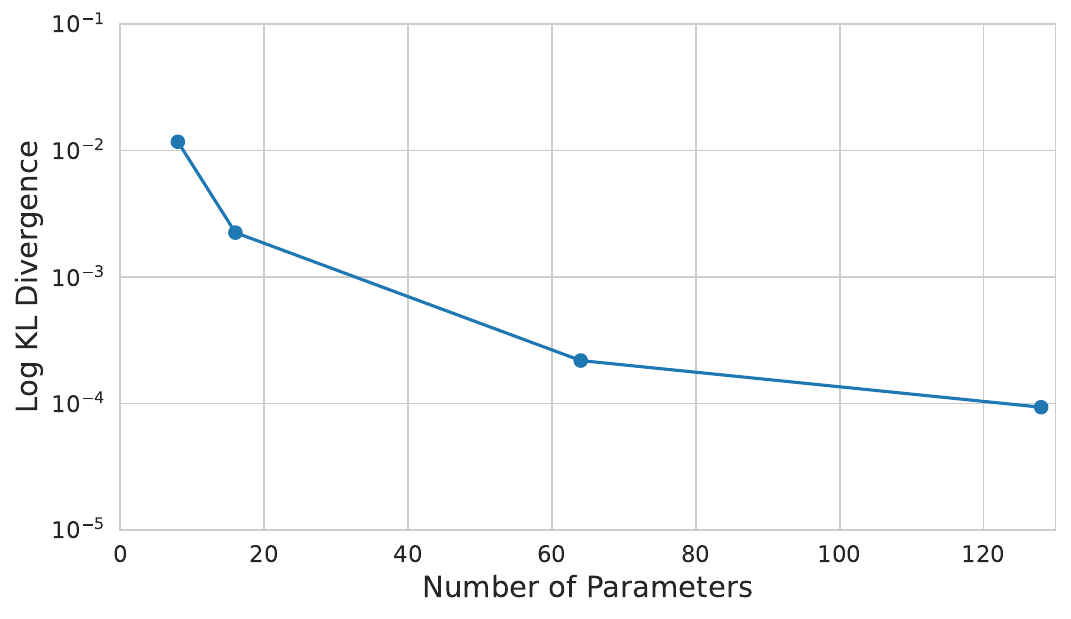}
        \caption{KLD vs. \# parameters}
    \end{subfigure}
    \caption{Model fit for mixture of beta distribution. a) Density plot for a $64$ term Fourier basis density model (best viewed on screen). b) The fit improves with increasing number of parameters.}
     \label{fig:mix_betas}
\end{figure}

\begin{figure}[t] 
    \centering
    \begin{subfigure}[t]{0.48\linewidth}
        \centering
        \includegraphics[width = \linewidth]{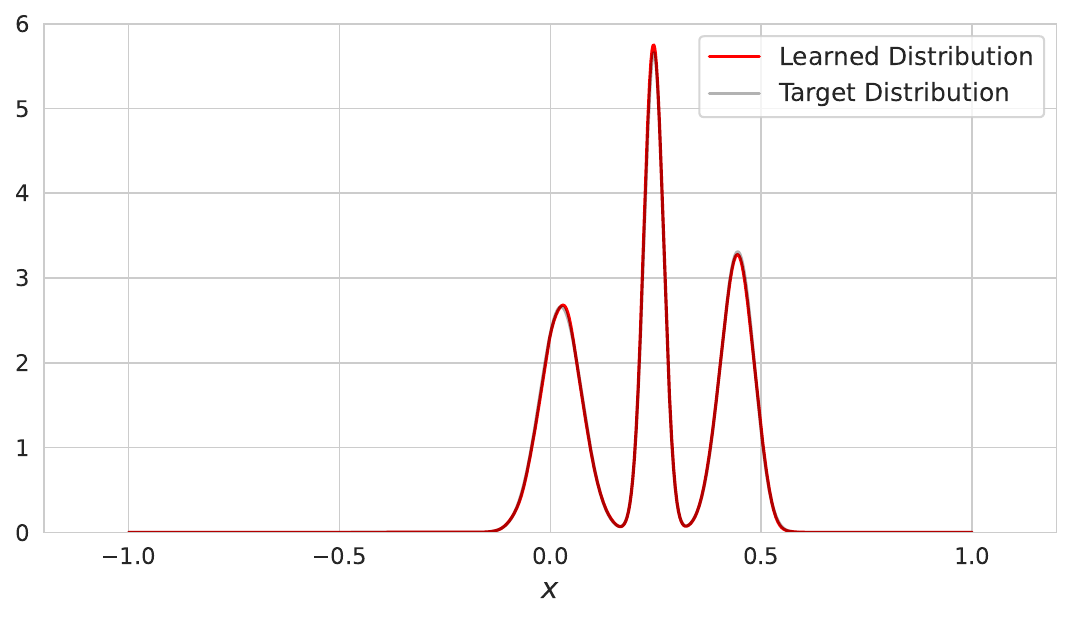}
        \caption{density fit, $N=64$}
    \end{subfigure}%
    \begin{subfigure}[t]{0.48\linewidth}
        \centering
        \includegraphics[width = \linewidth]{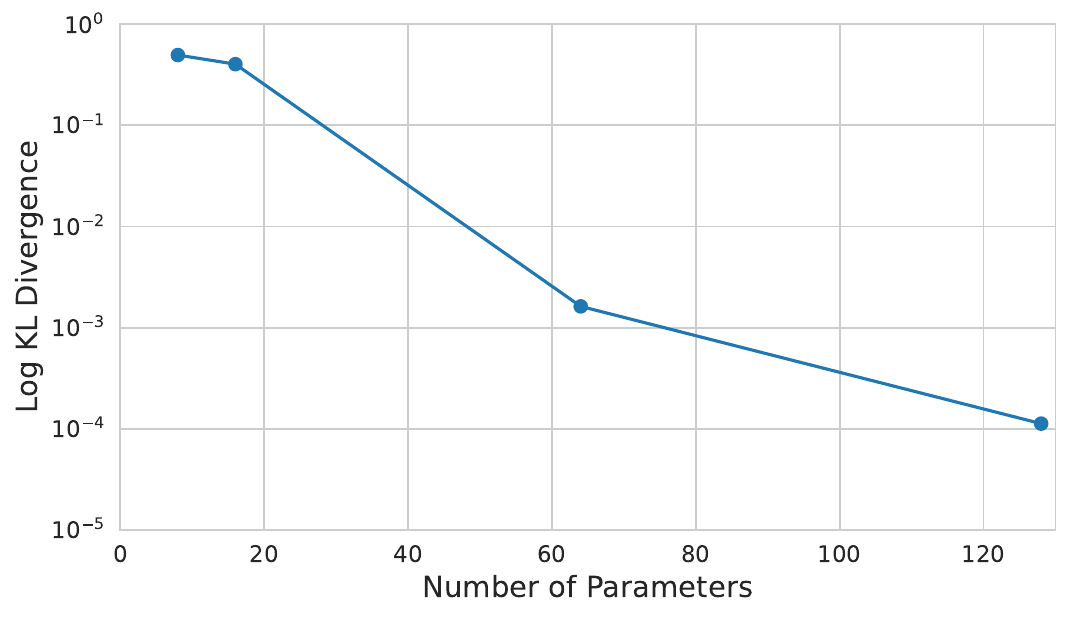}
        \caption{KLD vs. \# parameters}
    \end{subfigure}
    \caption{Model fit for mixture of logit-normals distribution.}
    \label{fig:mix_logitnormals}
\end{figure}

\begin{figure}[t]
    \centering
    \begin{subfigure}[t]{0.48\linewidth}
        \centering
        \includegraphics[width = \linewidth]{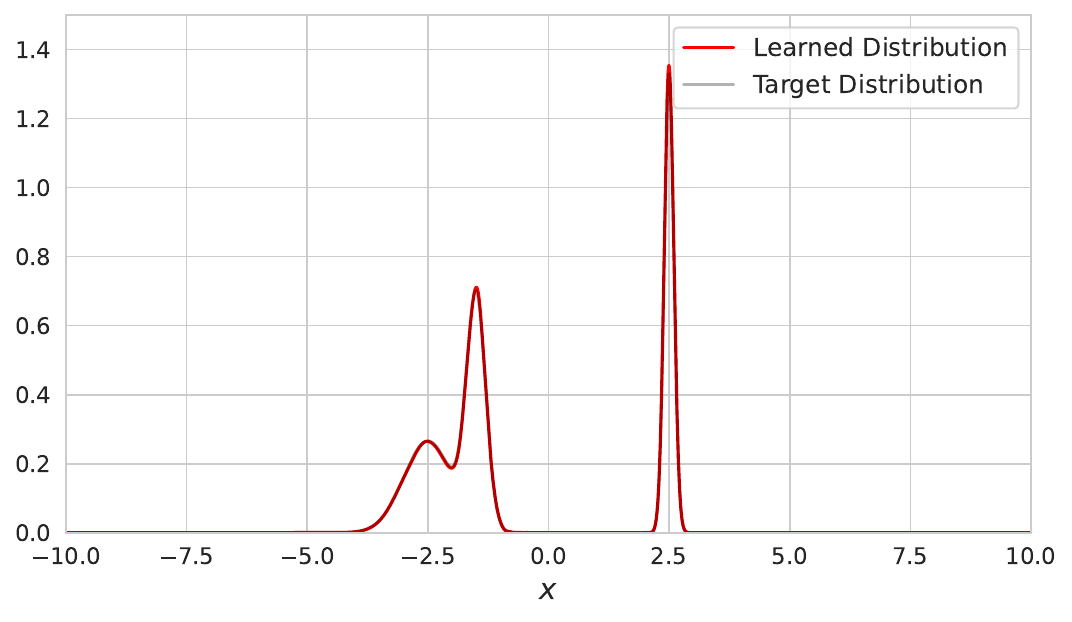}
        \caption{density fit, $N=64$}
    \end{subfigure}%
    \begin{subfigure}[t]{0.48\linewidth}
        \centering
        \includegraphics[width = \linewidth]{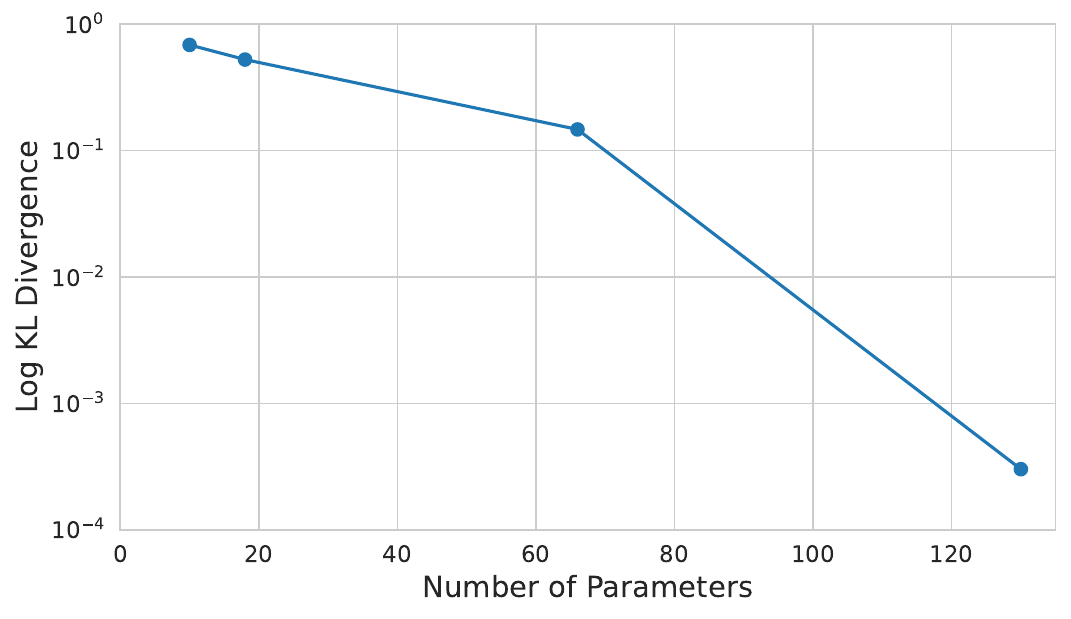}
        \caption{KLD vs. \# parameters}
    \end{subfigure}
    \caption{Model fit for mixture of 3 Gaussians. a) Density plot for a $64$ term Fourier basis density model (best viewed on screen). b) The fit improves with increasing number of parameters.}
     \label{fig:mix_gaussians_3}
\end{figure}

\begin{figure}[t]
    \centering
    \begin{subfigure}[t]{0.48\linewidth}
        \centering
        \includegraphics[width = \linewidth]{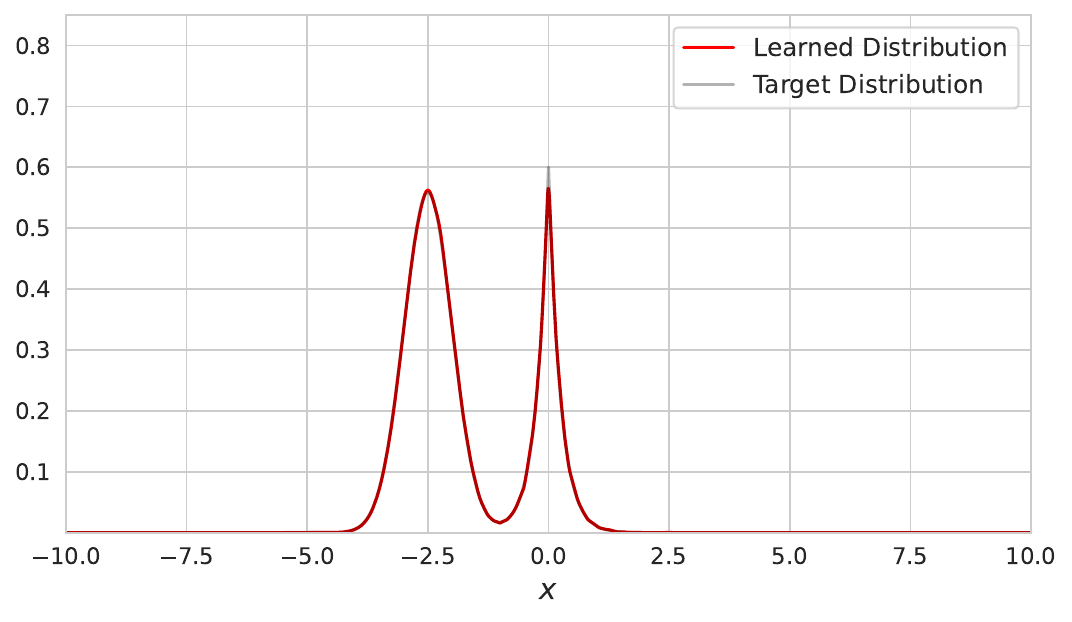}
        \caption{density fit, $N=64$}
\end{subfigure}%
    \begin{subfigure}[t]{0.48\linewidth}
        \centering
        \includegraphics[width = \linewidth]{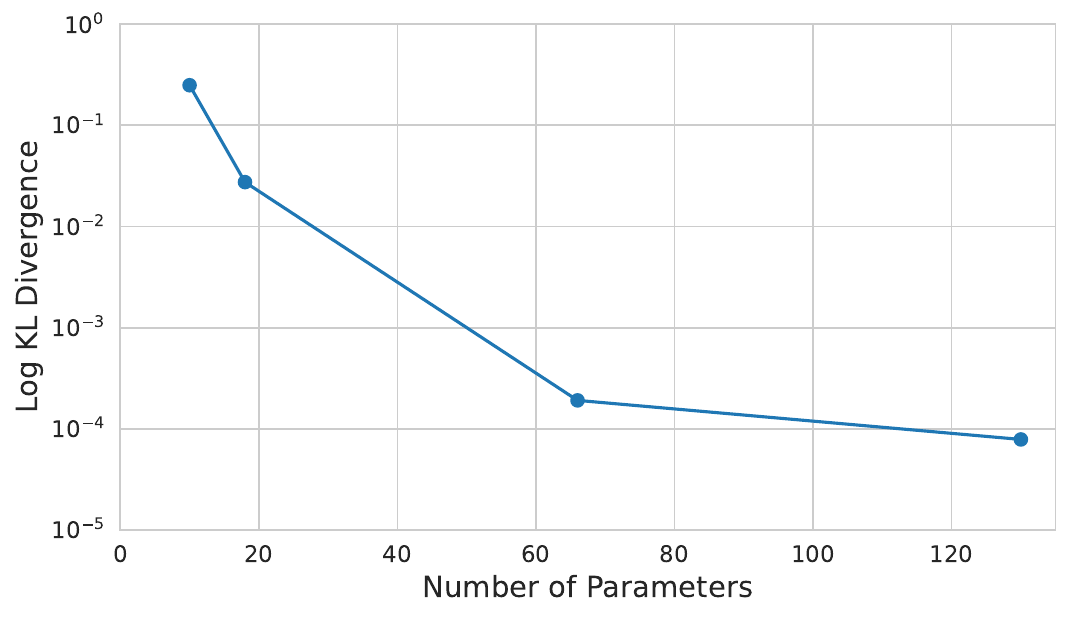}
        \caption{KLD vs. \# parameters}
    \end{subfigure}
    \caption{Model fit for mixture of Gaussians and Laplacians.}
     \label{fig:mix_gaussian_laplacian}
\end{figure}

\section{Experimental Evaluation}
In order to compare the density estimation and compression performance of our proposed model, we conducted a number of experiments using the CoDeX \cite{codex_github} library in the JAX framework \cite{jax2018github}.

\subsection{Experimental Setup}
We optimize the parameters of the density models by maximizing their log likelihood on samples from the target distributions, using the Adam optimizer \cite{kingma2017adam}, with a cosine decay schedule for the learning rate, initialized at $10^{-4}$. The models were trained for $500$ epochs of $500$ steps each, with a batch size of $128$ and a validation batch size of $2048$. After hyperparameter tuning, we found the regularization penalty $\gamma=10^{-6}$ to work well. For the DFP model, we consider neural network architectures with three hidden layers of $M$ units each, where $M \in \{5, 10, 20, 30\}$. Both models were trained using the same number of iterations, learning rate and optimizer. Further, we set the scale $s=1$ and offset $t=0$, unless otherwise specified.

As evaluation benchmarks, we experimented with univariate multi-modal distributions expressed as mixtures of Gaussian, beta, logit-normal, and Laplacian distributions. For the compression task, we use the banana distribution from \cite{balle2020nonlinear}.

\subsection{Impact of the Number of Frequency Terms on Model Fit}
Increasing the number of frequencies in our model affords us more expressive power, while also increasing the number of parameters. We studied this trade-off on 1) periodic distributions on the support $[-1, 1]$ and 2) distributions on $\mathbb{R}$.

For periodic distributions, we explored a mixture of two beta distributions and a mixture of logit-normal distributions. Results are reported in Fig. \ref{fig:mix_betas} and Fig. \ref{fig:mix_logitnormals}. In the left pane we display the fit qualitatively, while the right pane reports the KLD w.r.t. the true distribution as a function of the number of parameters (frequencies). We notice that the fit improves significantly with the number of parameters, until reaching a point of diminishing returns, where the addition of extra parameters does not produce a substantial difference.

Similarly, for the distributions with support on the real line, we considered a mixture of three Gaussian distributions and a mixture of Gaussian and Laplacian distributions (Fig. \ref{fig:mix_gaussians_3} and \ref{fig:mix_gaussian_laplacian}, respectively). We observe a significant decrease in KLD as the number of model of parameters is increased. By introducing the offset and scale terms in our model, we improve the quality of fit for asymmetric distributions.

\pgfplotsset{every tick label/.append style={font=\tiny}, label style={font=\tiny}, legend style={font=\tiny}}
\begin{figure}[t!]
    \centering
    \begin{subfigure}[t]{0.48\linewidth}
        \centering
        \includegraphics[width = 40mm]{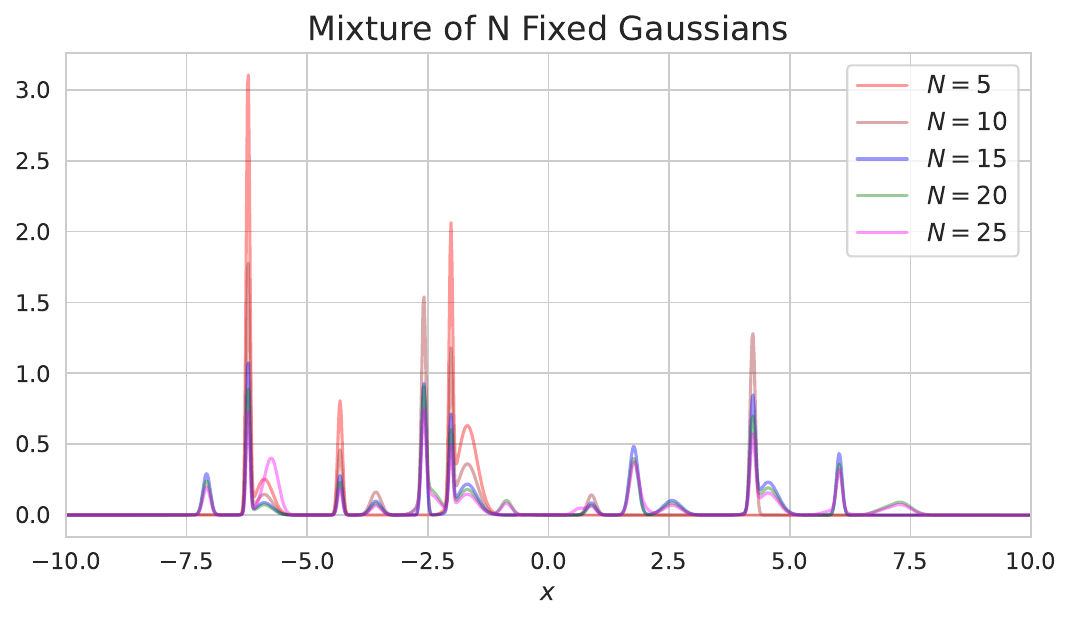}
        \caption{density fit}
    \end{subfigure}%
    \begin{subfigure}[t]{0.48\linewidth}
        \centering
        \includegraphics[width = 40mm]{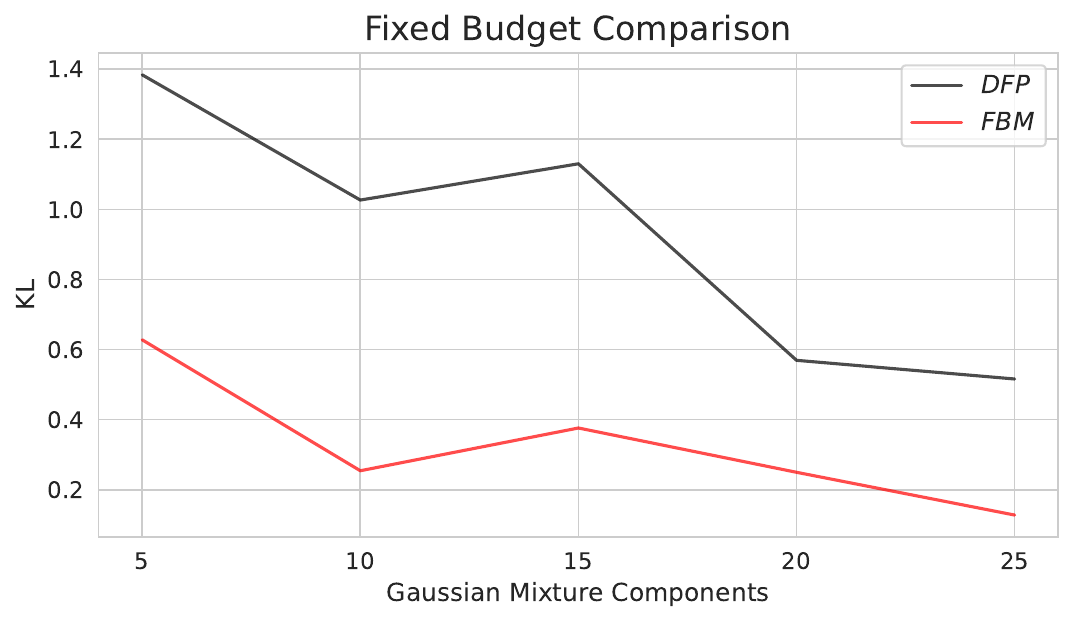}
        \caption{KLD vs. $K$}
    \end{subfigure}
    \caption{a) Model fit for mixture of $K$ Gaussians (best viewed on screen). b) KLD between model and target as a function of $K$, for deep factorized model (DFP) and Fourier basis density model (FBM) on a fixed parameter budget ($\sim 90$ parameters).}
    \label{fig:budget_results}
\end{figure}

\begin{figure}[t!]
    \centering
    \begin{subfigure}[t]{0.48\linewidth}
        \centering
        \includegraphics[width=\linewidth]{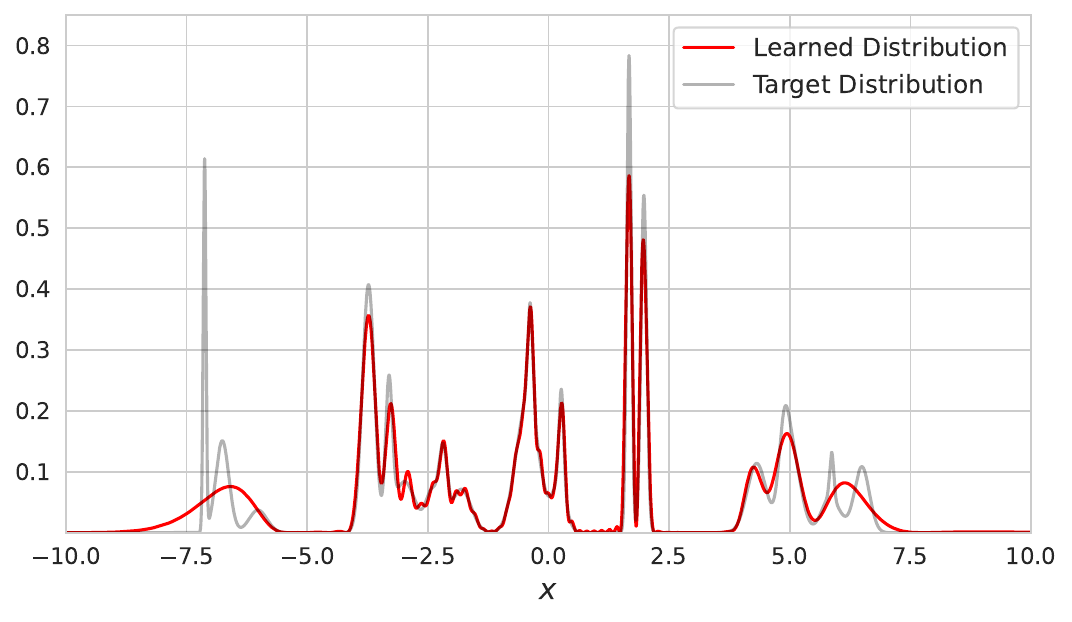}
        \caption{Fourier basis density model}
    \end{subfigure}%
    \begin{subfigure}[t]{0.48\linewidth}
        \centering
        \includegraphics[width=\linewidth]{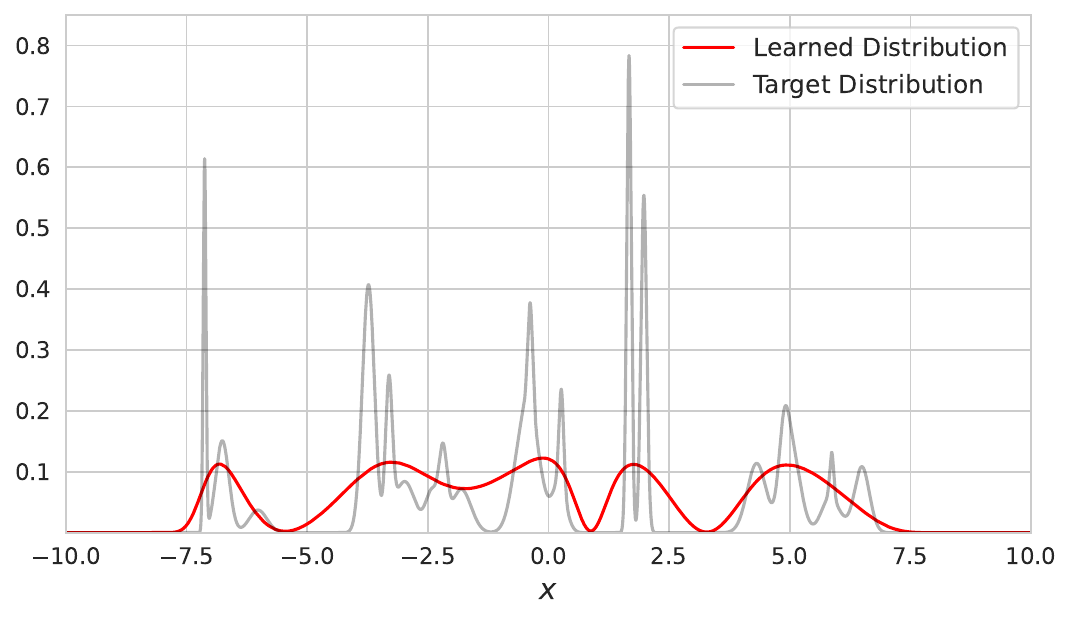}
        \caption{Deep factorized model}
    \end{subfigure}
    \caption{Model fit for mixture of $25$ Gaussians. a) Our proposed model captures most of the target distribution modes. b) In contrast, the Deep factorized model covers the modes, but also assigns a lot of probability mass to other regions.}
     \label{fig:fit_quality_25}
\end{figure}

\begin{figure}[ht]
\centering
\resizebox{0.48\textwidth}{!}{
\begin{tikzpicture}
\begin{axis}[
    height=4.5cm,
    width=6cm,
    ybar=1.5pt,
    bar width=8pt,
	ylabel=KL Divergence,
	xlabel=Number of Parameters,
	ymax=0.8,
	enlarge x limits=0.5,
	legend style={at={(0.8,1.)}, anchor=north, column sep=1pt},
	xtick=data,
    xticklabels={$\sim$ 90 params,$\sim$ 282 params},
    y axis line style={opacity=0},
]
\addplot 
	coordinates {(1, 0.3668) (2, 0.0541)};
\addplot 
	coordinates {(1, 0.5139) (2,0.5131)};
\addplot 
	coordinates {(1, 0.6239) (2,0.3531)};

\legend{Fourier basis density model, Deep factorized model, Gaussian mixture model}
\end{axis}
\end{tikzpicture}
}
\caption{Parameter efficiency evaluation. Our model provides a significantly better fit in comparison to the two baseline models of deep factorized model and Gaussian mixture model across two different parameter budget regimes. Accuracy of fit is measured in terms of KLD with respect to the target distribution, for a similar parameter budget for all three models. The target distribution is a heterogeneous mixture of Gaussian and Laplacian distributions.}
\label{fig:parameter_efficiency}
\end{figure}
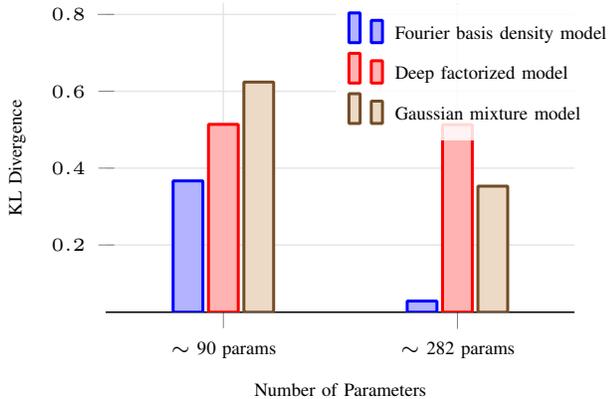

\begin{figure*}[t]
    \centering
    \begin{subfigure}[t]{0.32\linewidth}
        \centering
        \includegraphics[width = \linewidth]{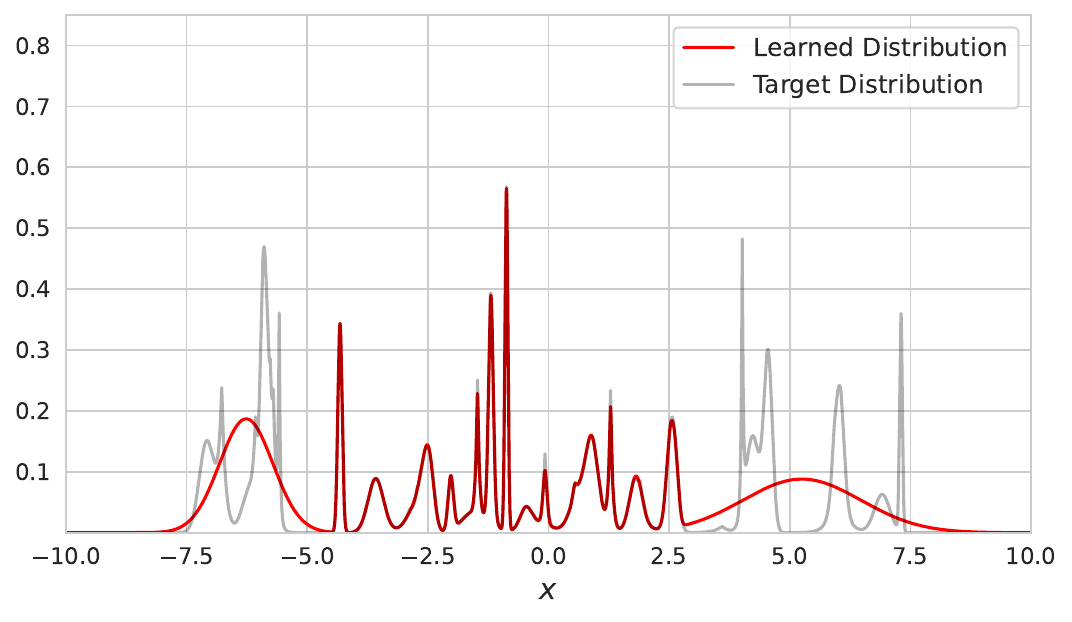}
        \caption{Gaussian mixture model}
    \end{subfigure}%
    \begin{subfigure}[t]{0.32\linewidth}
        \centering
        \includegraphics[width = \linewidth]{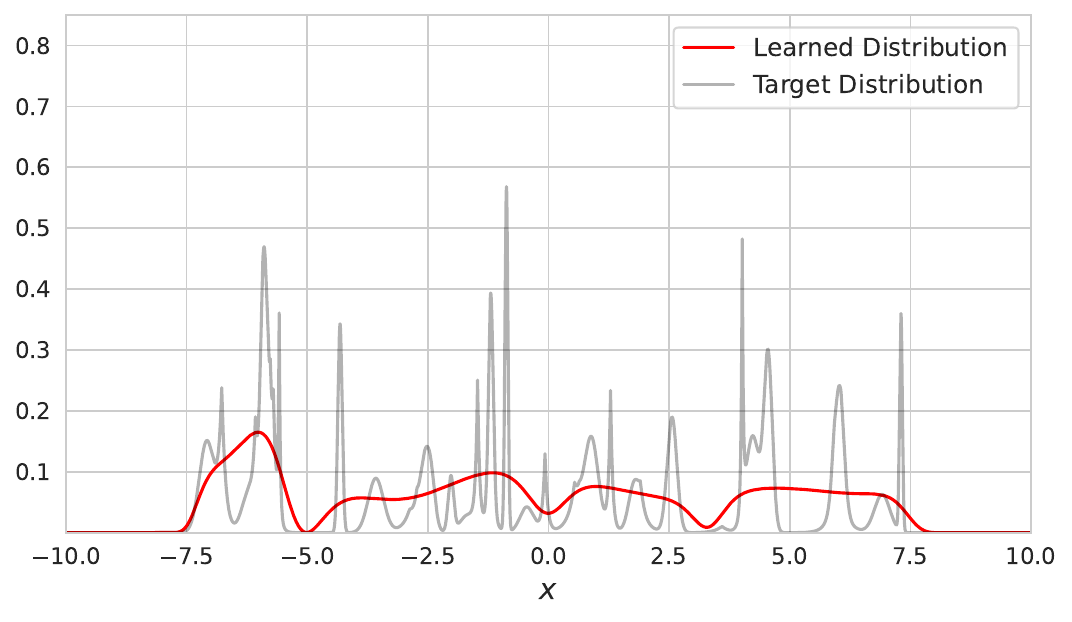}
        \caption{Deep factorized model}
    \end{subfigure}%
    \begin{subfigure}[t]{0.32\linewidth}
        \centering
        \includegraphics[width = \linewidth]{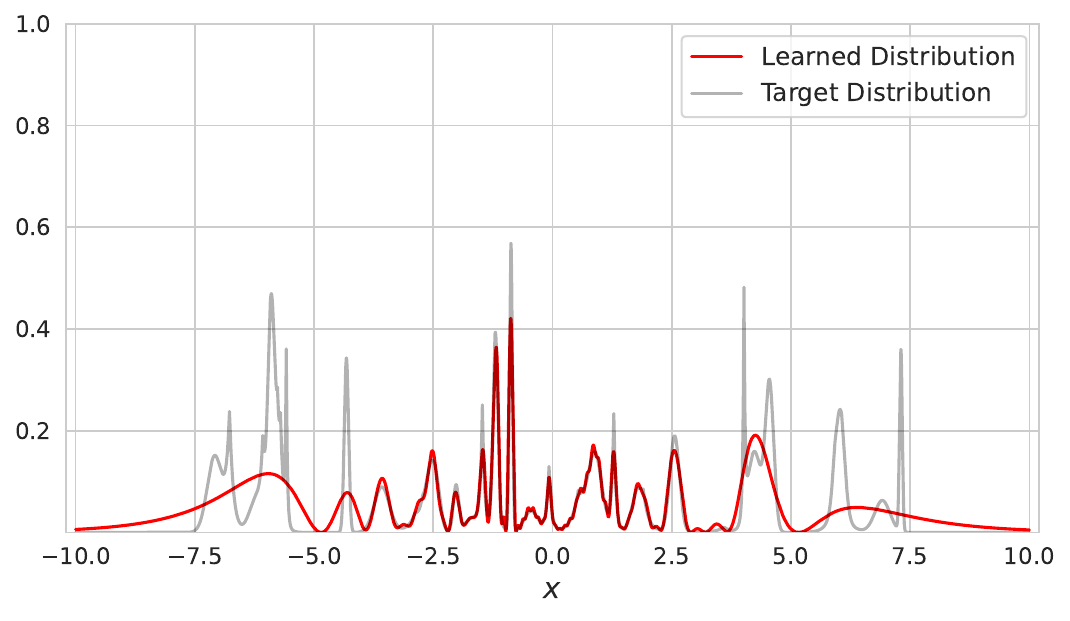}
        \caption{Fourier basis density model}
    \end{subfigure}
    
    \caption{Qualitative comparison of model fit for budget constrained models ($\sim 90$ parameters) for a multi-modal target distribution formed by a mixture of Laplacian and Gaussian distributions. Both the mixture of Gaussians as well as the Fourier basis density model fit most of the modes with precision, in comparison to the deep factorized model. Furthermore, the proposed model produces a better fit with the same amount of parameters with respect to the mixture of Gaussians, by fitting one extra mode around $x=4.0$ while sacrificing fit of the modes in the range $x=[-2.5, 0]$.}
     \label{fig:fixed_budget_high_dimensional}
\end{figure*}

\begin{figure*}[!htb]
    \centering
    \begin{subfigure}[t]{0.42\linewidth}
        \centering
        \includegraphics[width = \linewidth]{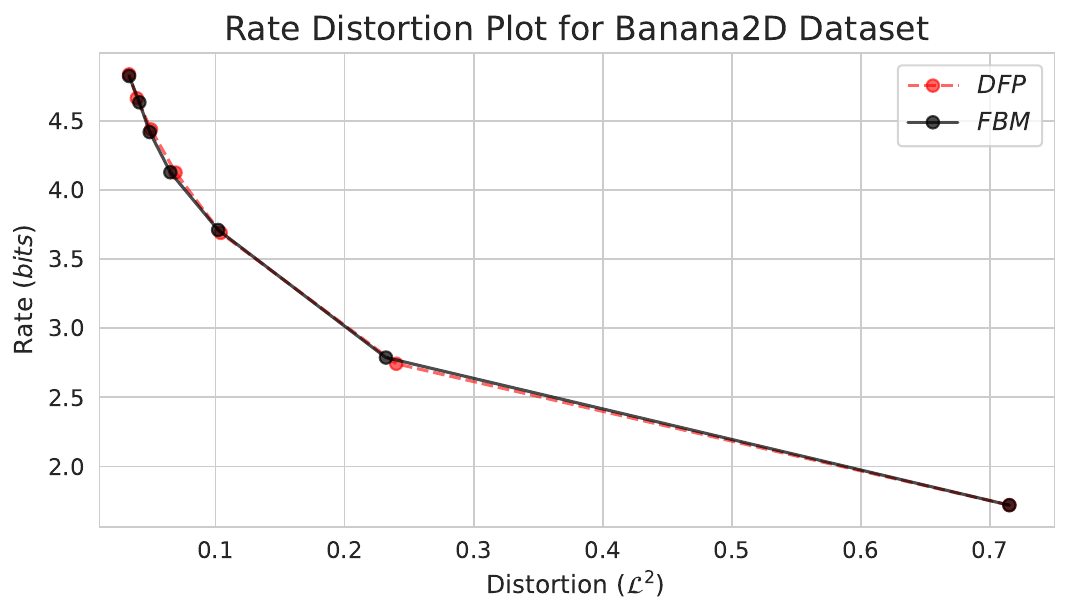}
        \caption{Rate--distortion performance}
    \end{subfigure}%
    \begin{subfigure}[t]{0.25\linewidth}
        \centering
        \includegraphics[width = \linewidth]{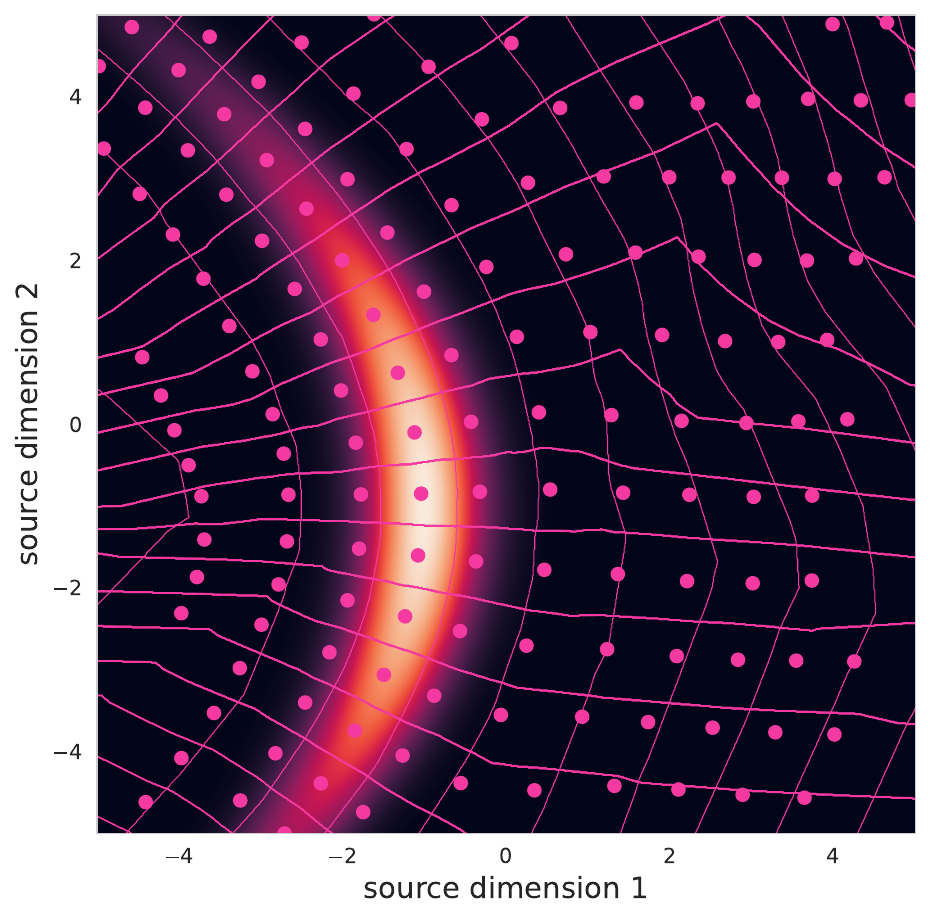}
        \caption{Fourier basis density model}
    \end{subfigure}%
    \begin{subfigure}[t]{0.25\linewidth}
        \centering
        \includegraphics[width = \linewidth]{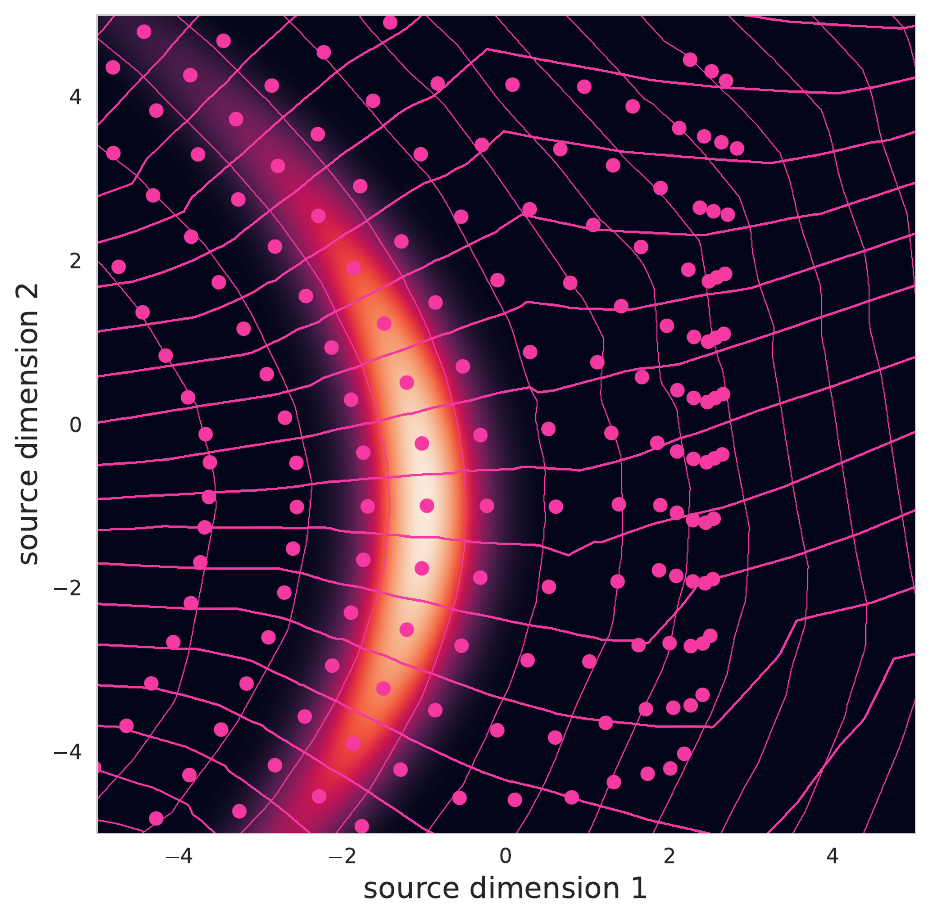}
        \caption{Deep factorized model}
    \end{subfigure}
    
    \caption{Rate--distortion comparison. a) R-D curves plotted by varying the trade-off parameter $\lambda$ over $\{ 1,5,10,15,20,25,30\}$. Our method (FBM) with $210$ parameters for the entropy coder slightly outperforms the deep factorized model (DFP) with $215$ parameters. b) and c): Quantization bins and bin centers learned by jointly optimizing rate--distortion using the Fourier basis entropy model and the deep factorized model, respectively, with a fixed $\lambda=10$. The results are quite comparable, with the exception of ``don't care'' regions off the main ridge, where the data distribution has few samples and the model behavior thus doesn't have an effect on the performance.}
     \label{fig:2d_banana}
\end{figure*}

\subsection{Multi-Modal Density Fit with a Fixed Parameter Budget}
We compared the performance of the Fourier basis density model against the deep factorized model at modeling highly multi-modal 1D distributions. We trained both models with a fixed parameter budget, and compared the KLD with respect to the target distribution. First, we evaluated a mixture of $K$ Gaussian distributions (for $K=\{5,10,15,20,25\}$) randomly located in the range $[-10, 10]$ with variance sampled inversely proportional to the number of components (Fig.~\ref{fig:budget_results}a). We observe that our model consistently obtains much lower KLD values compared to the DFP model for each value of $K$ (Fig.~\ref{fig:budget_results}b), validating the parameter efficiency of our model. Fig.~\ref{fig:fit_quality_25} shows that for the same number of parameters, the Fourier basis density model is able to capture the multi-modality of the distribution significantly better than the deep factorized model.

Next, we extended the experiment to a mixture of $20$ Laplace distributions and $20$ Gaussian distributions with mean values randomly located between $[-10, 10]$. Variances are sampled proportional to the number of components, and mixture weights are randomly sampled. Similar to the previous experiment, we compared the performance of the model with the deep factorized model, in terms of KLD with respect to the target distribution. Moreover, we also include a comparison with a Gaussian mixture model with a number of components such that the total number of parameters is equivalent for the three models. Results are visualized in Fig.~\ref{fig:parameter_efficiency} both for $\sim 90$ and $\sim 282$ parameters. We observe a significant gap in KLD between the models fitting the target distribution. Moreover, we show that as we overparameterize the models (with respect to the actual number of parameters of the target distribution), our model achieves remarkably lower KLD with an order of magnitude improvement compared to the other models. 

Fig.~\ref{fig:fixed_budget_high_dimensional} provides a qualitative comparison of the density fit achieved by various methods. We see that our model captures more modes of the target distribution in comparison to the other approaches, which explains its lower KLD. In contrast, the deep factorized model only roughly approximates the overall behavior of the highest modes in the distribution. Finally, even though the Gaussian mixture model produces a remarkable fit capturing some of the modes of the target distribution, the optimization problem is complex, and highly sensitive to the initialization of the parameters, leading to suboptimal fits for many of the modes.

\subsection{Lossy Compression of Banana Distribution}
A notable application of univariate density models is in nonlinear transform coding (NTC), where data is transformed to a latent space and each dimension of the latent representation is independently modeled and coded. To demonstrate the utility of our model, we evaluate our method on a compression task for the banana distribution introduced in \cite{balle2020nonlinear} (Fig. \ref{fig:2d_banana}). We follow \cite{balle2020nonlinear} closely and simply swap out the entropy model used during training. In brief, the model consists of an encoder, a decoder and an entropy model, which are jointly trained by minimizing the rate--distortion Lagrangian with respect to the model parameters $\theta$, i.e., the loss is
\begin{equation}
    \mathcal L_{\mathrm{compress}}(\theta) = R(\theta) + \lambda \, D(\theta),
\end{equation}
where $R$ is the rate and $D$ is the distortion. Further, $\lambda$ is a hyperparameter determining the desired trade-off. Here, $\theta$ includes the parameters of the entropy model as well as the non-linear transforms. We use squared error as a measure of distortion and a continuous and differentiable proxy for the discrete entropy as a measure of rate for the joint optimization during training. Once the model is trained, the latent space of the encoder can be uniformly scalar quantized for entropy coding. See \cite{balle2020nonlinear} for details.

We use eq. \eqref{eq:cdf_transformed} both to compute the probability within each quantization interval, in order to evaluate discrete entropy, as well as to obtain a model of the density convolved with a unit-width uniform distribution, for the differentiable proxy of entropy during training. For the encoder and decoder architectures, we use three-layer MLPs, with $50$ hidden units each, and leaky ReLU as the activation function. We consider a latent space dimension of $5$, learning rate of $10^{-3}$, batch size of $512$ samples, $200$ epochs of $2048$ steps, number of frequency terms $N=20$. 

Fig. \ref{fig:2d_banana}(a) plots the rate--distortion curves for the two approaches using deep factorized model (red) and Fourier basis density model (black). The curves are produced by varying the value of the trade-off parameter $\lambda$. As expected, the Fourier model achieves a comparable (or slightly better) trade-off in comparison to the deep factorized model, for all choices of~$\lambda$, and for an equivalent parameter budget. Figs.~\ref{fig:2d_banana}(b) and \ref{fig:2d_banana}(c) provide a qualitative comparison of the learned quantization bins and their representers for both models, with $\lambda=10$. Note that the bins far away from the mode see very few samples and therefore are not accurate. Further, the number and size of the quantization bins is a function of the trade-off ratio~$\lambda$. 

\section{Discussion}
We propose a novel univariate density model named Fourier basis density model, which is simple yet flexible and end-to-end trainable. Our experiments show that it provides a better fit for challenging multi-modal distributions in comparison to prevalent methods at a similar parameter budget, when trained with the same optimizer choices.
Moreover, the preliminary results obtained for the compression toy task show the effectiveness of our flexible density model in comparison to the deep factorized model as a building block for trainable end-to-end neural compression models.

\bibliographystyle{IEEEtran}
\bibliography{IEEEfull, References.bib}

\end{document}